\documentclass[10pt,twocolumn,letterpaper]{article}

\usepackage{wacv}              

\usepackage{graphicx}
\usepackage{amsmath}
\usepackage{amssymb}
\usepackage{booktabs}

\usepackage[pagebackref,breaklinks,colorlinks]{hyperref}

\usepackage[capitalize]{cleveref}
\crefname{section}{Sec.}{Secs.}
\Crefname{section}{Section}{Sections}
\Crefname{table}{Table}{Tables}
\crefname{table}{Tab.}{Tabs.}

\def\wacvPaperID{2187} 
\def\confName{WACV}
\def\confYear{2025}
\usepackage{amssymb}
\usepackage{amsmath,amsfonts}
\usepackage{algorithm}
\usepackage{algpseudocode}
\usepackage{array}
\usepackage{textcomp}
\usepackage{stfloats}
\usepackage{url}
\usepackage{verbatim}
\usepackage{graphicx}
\def\BibTeX{{\rm B\kern-.05em{\sc i\kern-.025em b}\kern-.08em
 T\kern-.1667em\lower.7ex\hbox{E}\kern-.125emX}}

\usepackage{subcaption}
\usepackage{multirow}
\usepackage{balance}
\usepackage{xspace}
\usepackage{xcolor}
\usepackage{booktabs}
\usepackage [english]{babel}
\usepackage [autostyle, english=american]{csquotes}
\MakeOuterQuote{"}
\usepackage{tabularx}
\usepackage{array}
\newcolumntype{C}{>{\centering\arraybackslash}X}
\usepackage{xstring}
\usepackage{enumerate}

\def\myVQVAE{\texttt{VQ-VAE}\@\xspace}
\def\myalgoname{\texttt{CE-VAE}\@\xspace}
\def\myGAN{\texttt{GAN}\@\xspace}
\def\myRbA{\texttt{RbA}\@\xspace}

\def\convtwod{\texttt{Conv2D}\@\xspace}
\def\convT{\texttt{TransposedConv}\@\xspace}

\def\resblock{\texttt{ResnetBlock}\xspace}
\def\upsampleblock{\texttt{UpSampleBlock}\@\xspace}

\newcommand{\tblfirst}[1]{\textbf{\textcolor{red}{#1}}}
\newcommand{\tblsecond}[1]{\textbf{\textcolor{blue}{#1}}}

\def\inputimg{\mathbf{I}}
\def\realcleanoutputimg{\inputimg^*}
\def\reconstructedimg{\widehat{\inputimg}^*}

\def\loss{\mathcal{L}}

\def\ie{\textit{i.e.}\@\xspace}
\def\eg{\textit{e.g.}\@\xspace}

\usepackage{graphicx}
\usepackage{etoolbox}

\newcommand{\insertfig}{\setcounter{figure}{0}\captionsetup{type=figure}\includegraphics[width=\linewidth]{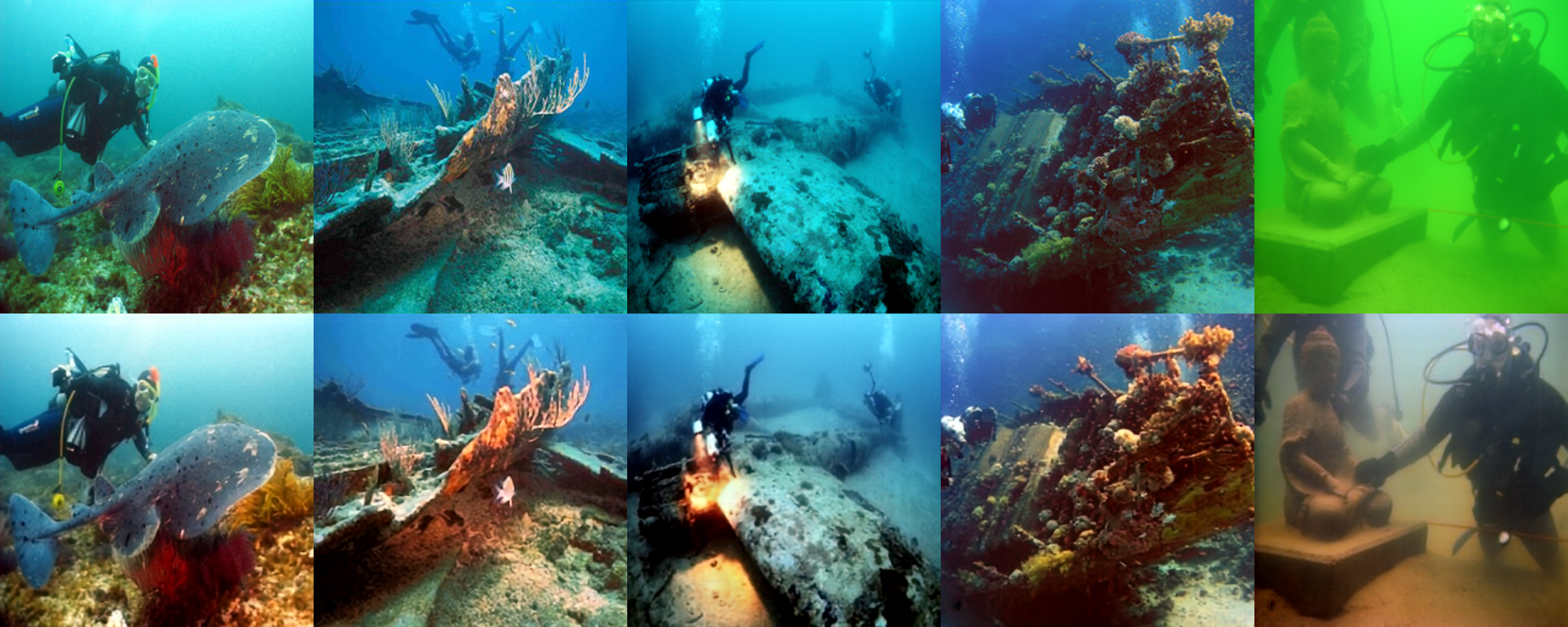}\captionof{figure}{Results of our approach for underwater compressed image reconstruction and enhancement. The first row shows the effects of water light refraction in underwater images introducing blurriness and cold greenish or bluish tone (among others) issues. The second row shows the enhanced results obtained by the proposed approach.}\label{fig:teaser}}

\renewcommand{\subsubsection}[1]{\textbf{#1}}

\makeatletter
\apptocmd{\@maketitle}{\centering\insertfig}{}{}

\makeatother

\begin{document}


\title{CE-VAE: Capsule Enhanced Variational AutoEncoder\\for Underwater Image Enhancement}

\author{Rita Pucci\textsuperscript{*}\\
Naturalis Biodiversity Center\\
Leiden, Netherlands (NL)\\
{\tt\small rita.pucci@naturalis.nl}
\and
Niki Martinel\textsuperscript{*}\\
University of Udine\\
Udine, Italy (IT)\\
{\tt\small niki.martinel@uniud.it}
}
\maketitle

\begingroup\renewcommand\thefootnote{*}
\footnotetext{Equal contribution}
\endgroup


\begin{abstract}
Unmanned underwater image analysis for marine monitoring faces two key challenges: (i) degraded image quality due to light attenuation and (ii) hardware storage constraints limiting high-resolution image collection.
Existing methods primarily address image enhancement with approaches that hinge on storing the full-size input.
In contrast, we introduce the Capsule Enhanced Variational AutoEncoder (\myalgoname), a novel architecture designed to efficiently compress and enhance degraded underwater images.
Our attention-aware image encoder can project the input image onto a latent space representation while being able to run online on a remote device.
The only information that needs to be stored on the device or sent to a beacon is a compressed representation. There is a dual-decoder module that performs offline, full-size enhanced image generation. One branch reconstructs spatial details from the compressed latent space, while the second branch utilizes a capsule-clustering layer to capture entity-level structures and complex spatial relationships.
This parallel decoding strategy enables the model to balance fine-detail preservation with context-aware enhancements.
\myalgoname achieves state-of-the-art performance in underwater image enhancement on six benchmark datasets, providing up to $3\times$ higher compression efficiency than existing approaches.
Code available at \url{https://github.com/iN1k1/ce-vae-underwater-image-enhancement}.
\end{abstract}

\section{Introduction}
Marine exploration is crucial for monitoring and protecting underwater ecosystems, with image analysis providing essential data for scientific research and environmental conservation. Recent advancements in unmanned and autonomous visual sensing systems have enhanced our ability to capture environmental data, enabling researchers to monitor~\cite{shkurti2012multi}, explore~\cite{whitcomb2000advances}, and analyze~\cite{bingham2010robotic} ocean depths while minimizing human risk.

However, underwater imagery poses significant challenges, such as severe color distortion and loss of detail due to light absorption, resulting in hazy images with greenish or bluish tinges (Fig.~\ref{fig:teaser}, first row). Addressing these degradation issues is critical for various marine applications, including ecological studies, underwater archaeology, and marine resource management.

In addition to image quality challenges, hardware storage limitations are a significant constraint for unmanned devices deployed in long-duration missions. These systems must operate autonomously for extended periods with limited capacity for storing high-resolution imagery~\cite{10678141,10639339}. Efficient image compression becomes essential to allow longer data collection campaigns without sacrificing the ability to perform high-quality image reconstruction and analysis offline.

Addressing both image degradation and storage efficiency issues is of paramount importance for autonomous marine monitoring systems, yet our community has mostly focused on the former problem.
Existing image enhancement methods can be categorized into: (i) traditional image processing techniques and (ii) machine learning-based methods. The former, including non-physics-based~\cite{li2016underwater,ghani2015underwater} and physics-based~\cite{han2017active, neumann2018fast,hu2021underwater} approaches, often lack generalization across diverse underwater environments. The latter~\cite{islam2020fast,park2019adaptive,hu2021underwater,fabbri2018enhancing,zhang2021dugan,zhu2017unpaired,islam2020simultaneous,guo2019underwater} offer superior generalization but are computationally intensive, limiting their integration into autonomous systems.

We introduce the Capsule Enhanced Variational AutoEncoder (\myalgoname), a novel architecture that synergizes the generative power of variational autoencoders with the strengths of capsule networks in capturing high-level image semantics ~\cite{sabour2017dynamic,pucci2021fixed,pucci2020deep,deng2018hyperspectral,pucci2021collaborative,pucci2022pro}.
\myalgoname comprises an encoder, a novel capsule layer, and a dual-decoder module, carefully designed to tackle the complex challenges of underwater image enhancement.

The novel attention-aware online encoder is designed to project the input image onto a highly compact low-dimensional latent representation.
This allows us to (i) achieve efficient storage of underwater imagery while (ii) also forcing the model to learn a compact, informative representation of the image, ensuring that irrelevant information and noise are discarded.

For offline full-size image enhancement, our architecture introduces a dual-decoder module, each designed with a specific role to ensure high-quality image reconstruction.
The first decoder reconstructs the enhanced image from the compressed latent space, ensuring the preservation of fine spatial details without requiring the full-size input. 
The second decoder leverages high-level features captured by a capsule layer, which provides a more abstract understanding of the scene.
The use of capsules is motivated by their ability to capture spatial hierarchies and part-whole relationships, making them robust to distortions in underwater imagery.
Capsules also dynamically cluster similar features, enhancing the model’s generalization across diverse underwater environments.
This dual-decoder design balances the retention of fine details with the ability to model complex structures, leading to superior image enhancement.

The key contributions of this work are:
\vspace{-1em}
\begin{itemize}
  \setlength{\itemsep}{0pt}
  \setlength{\parskip}{0pt}
  \setlength{\parsep}{0pt}
\item A novel attention-aware encoder that projects input images onto a highly compressed latent space, enabling efficient storage and real-time processing for underwater image enhancement.
\item A dual-decoder architecture that reconstructs enhanced images by leveraging both compressed latent representations and high-level capsule features, balancing spatial detail preservation with context-aware reconstruction.
\item State-of-the-art performance on six benchmark datasets, achieving superior image quality and generalization while offering $3\times$ improved storage efficiency by eliminating the need to store full-size input images.
\end{itemize}

\section{Related Works}
\label{sec:rw}
\subsubsection{Traditional methods} focus on the estimation of global background and water light transmission to perform image enhancement.
In~\cite{bazeille2006automatic,8058463}, independent image processing steps have been proposed to correct non-uniform illumination, suppress noise, enhance contrast, and adjust colors.
Other methods introduced edge detection operations to implement object-edge preservation during filtering operations for color enhancement~\cite{lu2013underwater}.
In~~\cite{li2016single}, it has been observed that the image channels are affected differently by the disruption of light: red colors are lost after a few meters from the surface while green and blue are more persistent.
These differences introduced enhancement methods that act differently on each color channel and sacrifice generalization in favour of ad-hoc filters based on environmental parameters~\cite{park2017enhancing, 9744022}.
Other approaches estimated the global background light parameters~\cite{park2017enhancing, peng2017underwater} to apply specific color corrections (\ie, to reduce the blueish and greenish effects).
These models use the principles of light and color physics to account for various underwater conditions. Despite being more accurate, their application is limited due to the challenges of obtaining all the necessary variables that impact underwater footage.
Efforts have been made to improve the estimation of the global background light~\cite{akkaynak2019sea} at the cost of increasing algorithm complexity and overfitting experimental data with poor generalization on new test data.

\begin{figure*}[th]
  \centering
  \includegraphics[width=\linewidth]{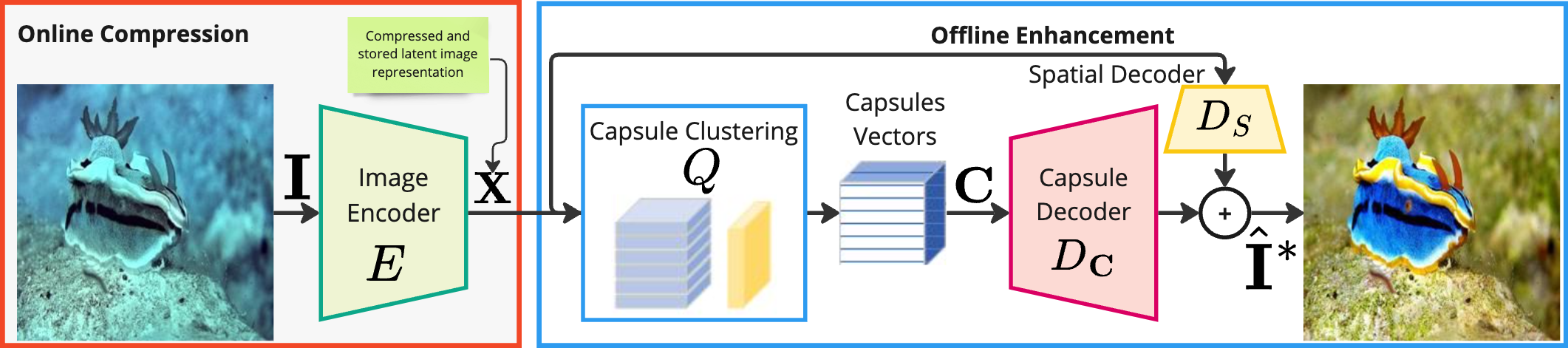}
  \caption{Proposed \myalgoname architecture with the new capsule vector latent space clusterization mechanism.}
  \label{fig:system_pipeline}
\end{figure*}

\subsubsection{Machine learning-based methods} for underwater image enhancement made extensive use of a U-Net-like structure~\cite{ronneberger2015u} to enhance the input image while preserving the spatial information and relationship between objects.
Skip connections are often used to propagate the raw inputs to the final layers to preserve spatial relationships~\cite{8917818,9986534} also with special attention and pooling layers~\cite{qiao2022adaptive}.
Other methods explored the emerging application of Transformers via channel-wise and spatial-wise attention layers~\cite{peng2023ushape-lsuidataset} or through customized transformer blocks leveraging both the frequency and the spatial-domains as self-attention inputs~\cite{khan2024spectroformer}.
Generative Adversarial Networks (\myGAN{}s) training schemes have also been explored for the task~\cite{guo2019underwater} along with 
approaches improving the information transfer between the encoder and decoder via multiscale dense blocks~\cite{islam2020fast} or hierarchical attentions modules~\cite{han2023fe}.

Our approach falls in the latter category. In contrast with such methods, we propose a novel architecture that removes the need for skip connections between the raw input and decoder layers.
Our encoder projects the full-size input image into a highly compact, low-dimensional latent space that captures all relevant information for both enhancement and reconstruction.
The dual-decoder module operates exclusively on this latent representation, fully independent of the full-size raw input.
This design allows for real-time feature extraction during data collection, enabling efficient storage by retaining only the latent compressed representation, which is crucial for resource-constrained environments.

\section{Proposed Method}
\label{sec:method}
Figure \ref{fig:system_pipeline} illustrates our architecture, composed of two main phases.
The online compression phase features an image encoder ($E$) that models the degraded input image $\inputimg \in \mathbb{R}^{3\times H \times W}$ and compresses it into a latent feature space carrying relevant information for enhancement.
The compressed representation can then be stored and later used in the offline enhancement phase.
This includes a capsule clustering module ($Q$), capturing entity-level features, followed by the capsule decoder ($D_\mathbf{C}$) and the spatial decoder ($D_S$) that jointly collaborate to generate the full-size enhanced image,~\ie, $\reconstructedimg \in \mathbb{R}^{3\times H \times W}$.

\subsection{Image Encoder ($E$)}
\label{sec:method:encoding}
Our encoder architecture is designed to extract a compact yet informative latent representation while preserving crucial spatial information. The design follows a hierarchical structure that balances computational efficiency with feature richness.

We begin by computing $\mathbf{H}_0 = \texttt{Conv2D}_{3\times3}(\inputimg)$ to capture low-level features such as edges, textures, and color variations, which are often distorted in underwater environments.
This is followed by $N$ encoding blocks, each comprising a residual block and a self-attention mechanism with skip connections.
The residual block computes $\mathbf{H}_{l}^{\text{res}}$ where $l\in [1,N]$
\begin{equation}
\mathbf{H}_{l}^{\text{res}} = \resblock(\mathbf{H}_{l-1})\in \mathbb{R}^{C_{l} \times H_{l} \times W_{l}}
\end{equation}
 ensuring effective information propagation through deeper layers for preserving and enhancing subtle underwater textures and colors, while mitigating vanishing gradients.

The subsequent self-attention mechanism further refines the extracted features to get
\begin{equation}
\mathbf{H}_l = \mathbf{H}_l^{\text{res}} + \texttt{SelfAttention}(\mathbf{H}_l^{\text{res}})
\end{equation}
The attention block allows the model to focus on salient regions in the feature map, which is particularly important for underwater images where certain areas may be more affected by scattering, absorption, or color distortion than others.

Between every two encoding blocks, we also add a $\texttt{Conv2D}$ to half feature resolution spatial dimensions, optimizing computational efficiency while allowing the model to capture high-level abstract features.
The balance between resolution and abstraction is essential for processing large underwater images efficiently while preserving critical information about the global color cast and lighting conditions.

At the output of the $N$ encoder blocks, we add a convolutional block-computing
\begin{equation}
\mathbf{X} = \texttt{Conv2D}_{3\times 3}(\texttt{Swish}(\texttt{BN}(\mathbf{H}_{N}))) \in \mathbb{R}^{C_X \times H_X \times W_X}
\end{equation}
that refines the learned representation and produces the desired compact yet informative latent space.

\subsection{Capsule Clustering ($Q$)}
\label{sec:method:cvq}
\begin{figure}
  \centering
  \includegraphics[width=\linewidth]{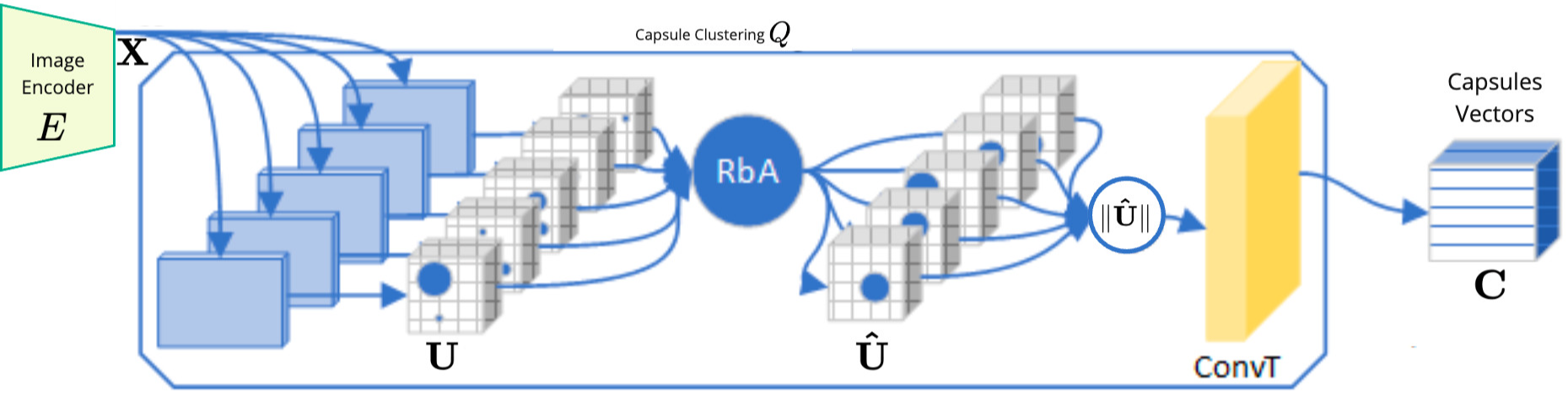}
  \caption{Proposed capsule vector clustering approach.
  It consists of a capsule layer and a convolutional transpose layer.
  The capsules extract $\mathbf{U}$ features which are clusterized by the \myRbA{ }procedure, to obtain $\mathbf{\hat{U}}$. We aggregate the matrices and upsample them by a transposed convolution layer.}
  \label{fig:CV}
\end{figure}

Following the encoder, the capsule layer processes the compressed latent representation to model entity-level relationships within the image (see \figurename~\ref{fig:CV}). Given $L$ as the first capsule layer and $L+1$ as the consecutive one. First, $\beta_L$ parallel convolutional layers (\convtwod) process the encoder output $\mathbf{X}$, generating a tensor $\mathbf{U} \in \mathbb{R}^{\beta_L \times C_{\mathbf{U}} \times H_{\mathbf{U}} \times W_{\mathbf{U}}}$, where $C_{\mathbf{U}}$ is the number of channels and $H_{\mathbf{U}}, W_{\mathbf{U}}$ are the spatial dimensions. For each spatial location, $\mathbf{u}_i \in \mathbb{R}^{C_{\mathbf{U}}}$ represents the output of capsule $i$ at level $L$. Capsule $j\in[0,k]$ at the next level, $L+1$, receives information from all capsules at $L$, and computes the affine transformation of $\mathbf{u}_i$:

\begin{equation}
    \hat{\mathbf{u}}_{j|i} = \mathbf{W}_{ij} \mathbf{u}_i
\end{equation}

where $\mathbf{W}_{ij} \in \mathbb{R}^{C_{\mathbf{U}} \times C_{\hat{\mathbf{U}}}}$ is a weight matrix defining how capsule $i$ contributes to capsule $j$. The vector $\hat{\mathbf{u}}_{j|i}$ estimates the relevance of capsule $i$ for activating capsule $j$. Since not all parent capsules are equally important for higher-level entities, we apply a coupling coefficient $c_{ij}$ to weigh their contributions. The softmax function is used to compute $c_{ij}$:

\begin{equation}
    c_{ij} = \frac{\exp(b_{ij})}{\sum_{k}\exp(b_{ik})}
\end{equation}

where $b_{ij}$ is iteratively updated during the Routing-by-Agreement (\myRbA) process. The next step is to compute a weighted sum of the prediction vectors:

\begin{equation}
    \mathbf{s}_j = \sum_{i} c_{ij} \hat{\mathbf{u}}_{j|i}
\end{equation}

The squashing function is then applied to obtain the activity vector $\mathbf{v}_j$, representing the likelihood of entity presence:

\begin{equation}
    \mathbf{v}_j = \frac{\Vert\mathbf{s}_j\Vert^2}{1+\Vert\mathbf{s}_j\Vert^2}\frac{\mathbf{s}_j}{\Vert\mathbf{s}_j\Vert}
\end{equation}

Finally, $b_{ij}$ is updated based on the agreement between $\mathbf{v}_j$ and $\hat{\mathbf{u}}_{j|i}$, where agreement strengthens the contribution of capsule $i$ to capsule $j$.

The activity vector $\mathbf{v}_j$ carries information about the presence of an entity using a probability representation that removes information about the precise (pixel-level) spatial location of each relevant object.
Such information, which is very relevant for precise enhanced image reconstruction, is however contained in the related prediction vectors.
To bring the spatial information to the next layers, while preserving the information about entities, we first weight every prediction vector from level $L$ (\ie, $\hat{\mathbf{u}}_{j\vert i}$) by the corresponding coupling coefficient $c_{ij}$ (estimated at the last iteration of the \myRbA procedure) to collectively obtain $\hat{\mathbf{U}} \in \mathbb{R}^{\beta_{L+1} \times C_{\hat{\mathbf{U}}}\times H_{\hat{\mathbf{U}}} \times W_{\hat{\mathbf{U}}}}$.
Then, entity presence at a specific location is then captured through the $\ell_2$-norm computation over $\beta_{L+1}$.
Finally, a \convT layer is exploited to obtain the capsule vectors $\mathbf{C} \in \mathbb{R}^{C_{\mathbf{C}} \times H_{\mathbf{U}} \times W_{\mathbf{U}}}$.

The motivation for using the capsule layer in this setting lies in its ability to capture spatial hierarchies and model part-to-whole relationships, which are key to representing entities within degraded underwater images.
CNNs struggle to maintain precise entity-based information across layers, often losing finer details necessary for effective reconstruction.
By employing the capsule networks mechanism while preserving the spatial pixel relationships via $\hat{\mathbf{U}}$, \myalgoname maintains both spatial and entity-level information. 
This is particularly important in underwater imagery, where preserving the structure of small objects and understanding spatial relationships is essential for accurate enhancement. 

\subsection{Decoding}
To reconstruct the enhanced image, we introduce two parallel decoders: the capsule decoder ($D_\mathbf{C}$) and the spatial decoder ($D_S$).

$D_{\mathbf{C}}$ reconstructs the image leveraging the information about the presence of entities identified by the capsules vectors,~\ie, $\mathbf{C}$.
It increases the input spatial feature map resolution to produce $\reconstructedimg_{D_\mathbf{C}}\in \mathbb{R}^{3\times H\times W}$ by a sequence of 4 blocks, each consisting of a \resblock and an \upsampleblock ~\cite{esser2021taming}. 
This decoder works on an input that contains information about the presence of entities in the image but might not contain precise (\textit{i.e.}, to pixel-level) information about their displacement.
Since the enhancement must generate an output that preserves all the spatial details but removes the effects of underwater degradation, this information would be very relevant for reconstruction.

To mitigate such a limitation, we introduced the spatial decoder ($D_S$).
This takes as input the low-resolution feature map, $\mathbf{X}$, that preserves all the spatial details about the input, and gradually increases its resolution to match the input image size, with a process that resembles image super-resolution works.
Similarly to $D_\mathbf{C}$, such a module is composed of 4 blocks, each consisting of a \convT and a \resblock, that emit $\reconstructedimg_{D_S} \in \mathbb{R}^{3\times H\times W}$. 

The decoders produce the model output $\reconstructedimg = \reconstructedimg_{D_\mathbf{C}} + \reconstructedimg_{D_S}$.

\subsection{Optimization Objective}
The proposed architecture utilizes only the compressed latent representation extracted by $E$ for both decoders $D_C$ and $D_S$. This design enables edge computation in $E$ of $\mathbf{X}$, and subsequent offline reconstruction via $D_C$ and $D_S$. This approach effectively functions as a data compression method, facilitating extended underwater acquisition campaigns. However, maximizing compression efficiency necessitates learning a highly informative latent image representation. To achieve this, we introduce a composite loss function that encapsulates the essential aspects of underwater image enhancement: preservation of the spatial structure, improved color perception with artifact suppression, and overall image realism.

\subsubsection{Reconstruction Loss.}
To ensure spatial coherence between the noise-free ground truth (\ie, $\realcleanoutputimg$) and reconstructed image (\ie, $\reconstructedimg$), we compute:
\begin{equation}
\loss_{rec}= | \realcleanoutputimg - \reconstructedimg |
\end{equation}
However, the model may generate blurry or overly smooth images, as the network tries to minimize the pixel-wise differences without necessarily capturing the high-level features and structures of the original image.

\subsubsection{Perceptual Loss.}
We employ the Learned Perceptual Image Patch Similarity (LPIPS) metric, which has been shown to correlate well with human judgments of image quality~\cite{7797130}.
This computes:
\begin{equation}
    \label{eq:lpips_loss}
    \loss_{lpips} = ||\phi(\realcleanoutputimg) - \phi(\reconstructedimg)||_2
\end{equation}
where $\phi(\cdot)$ denotes a pre-trained model extracting features relevant to human perception.

\subsubsection{Adversarial Loss.}
To further improve the realism of the generated images, we adopted an adversarial training procedure with a patch-based discriminator $\psi(\cdot)$~\cite{isola2017image}.
We follow the original formulation of~\cite{esser2021taming} to define
\begin{equation}
\loss_{GAN} = \lambda \left( log \psi(\realcleanoutputimg) + log(1 - \psi(\reconstructedimg)) \right) 
\end{equation}
where its contribution to the final objective is controlled by 
\begin{equation}
    \lambda = \frac{\nabla_{D_{\mathbf{C}}}(\mathcal{L}_{rec})}{\nabla_{D_{\mathbf{C}}}(\mathcal{L}_{GAN}) + \delta}
\label{lambda}
\end{equation}
$\nabla_{D_{\mathbf{C}}}(\cdot)$ is the gradient of its input at the last layer of $D_{\mathbf{C}}$, and $\delta$ is used for numerical stability~\cite{esser2021taming}.

\subsubsection{Structural Similarity Loss.}
To address the structural distortions common in underwater image degradation, we consider the Structural Similarity Index Measure (SSIM)~\cite{wang2004ssim} loss function:
\begin{equation}
\label{eq:ssim_loss}
\loss_{SSIM} = \frac{1}{M} \sum_{i=1}^{M} 
\frac{2 \mu_{\realcleanoutputimg_i} \mu_{\reconstructedimg_i} + \kappa_1}
{
\mu_{\realcleanoutputimg_i}^2 + \mu_{\reconstructedimg_i}^2 +
\kappa_1
} 
\frac{2 \sigma_{\realcleanoutputimg_i \reconstructedimg_i} + \kappa_2}
{
\sigma_{\realcleanoutputimg_i}^2 + \sigma_{\reconstructedimg_i}^2 + \kappa_2}
\end{equation}
where $\realcleanoutputimg_i$ and $\reconstructedimg_i$ are $11\times 11$ non-overlapping image patches, $\mu$ and $\sigma$ represent the mean and standard deviation operators.
$\kappa_1$ and $\kappa_2$ are small constants added for stability.

\subsubsection{Combined Loss.}
Our optimization objective is
\begin{equation}
\loss =  \loss_{rec} + \loss_{lpips} + \loss_{GAN} + \loss_{SSIM}
\end{equation}

\begin{table*}[t]
\centering
\scriptsize
\begin{tabularx}{\linewidth}{l C C C C C C C C C}
\toprule
                    & \multicolumn{3}{c}{LSUI-L400} & \multicolumn{3}{c}{EUVP} & \multicolumn{3}{c}{UFO-120} \\
                    \cmidrule(lr){2-4} \cmidrule(lr){5-7} \cmidrule(lr){8-10}
                    & PSNR $\uparrow$ & SSIM $\uparrow$ & LPIPS $\downarrow$ & PSNR $\uparrow$ & SSIM $\uparrow$ & LPIPS $\downarrow$ & PSNR $\uparrow$ & SSIM $\uparrow$ & LPIPS $\downarrow$ \\
\midrule
RGHS~\cite{huang2018rghs}                & 18.44 & 0.80 & 0.31 & 18.05 & 0.78 & 0.31 & 17.48 & 0.71 & 0.37 \\
UDCP~\cite{drews2016udcp}                & 13.24 & 0.56 & 0.39 & 14.52 & 0.59 & 0.35 & 14.50 & 0.55 & 0.42 \\
UIBLA~\cite{peng2017uibla}               & 17.75 & 0.72 & 0.36 & 18.95 & 0.74 & 0.33 & 17.04 & 0.64 & 0.40 \\
UGAN~\cite{fabbri2018enhancing}          & 19.40 & 0.77 & 0.37 & 20.98 & 0.83 & 0.31 & 19.92 & 0.73 & 0.38 \\
FUnIE-GAN~\cite{islam2020fast}           & -     & -    & -    & 23.53 & 0.84 & 0.26 & 23.09 & 0.76 & 0.32 \\
Cluie-Net~\cite{li2023cluienet}          & 18.71 & 0.78 & 0.33 & 18.90 & 0.78 & 0.30 & 18.43 & 0.72 & 0.36 \\
DeepSESR~\cite{islam2020simultaneous}    & -     & -    & -    & 24.22 & 0.85 & 0.25 & \tblsecond{23.38} & \tblsecond{0.78} & \tblsecond{0.29} \\
TWIN~\cite{liu2022twin}                  & 19.84 & 0.79 & 0.33 & 18.91 & 0.79 & 0.32 & 18.21 & 0.72 & 0.37 \\
UShape-Transformer~\cite{peng2023ushape-lsuidataset} & \tblsecond{23.02} & \tblsecond{0.82} & \tblsecond{0.29} & \tblsecond{27.59} & \tblsecond{0.88} & \tblsecond{0.23} & 22.82 & 0.77 & 0.33 \\
Spectroformer~\cite{khan2024spectroformer} & 20.09 & 0.79 & 0.32 & 18.70 & 0.79 & 0.32 & 18.03 & 0.71 & 0.37 \\
\myalgoname                              & \tblfirst{24.49} & \tblfirst{0.84} & \tblfirst{0.26} & \tblfirst{27.75} & \tblfirst{0.89} & \tblfirst{0.20} & \tblfirst{24.38} & \tblfirst{0.79} & \tblfirst{0.28} \\
\bottomrule
\end{tabularx}
\caption{Quantitative comparison of \myalgoname and state-of-the-art methods across the three considered full-reference datasets: LSUI-L400, EUVP, and UFO-120. ($\uparrow$ higher is better, $\downarrow$ lower is better). For each metric/dataset the best method is in red, second best in blue.}
\label{tab:full_reference_sota}
\vspace{0em}
\end{table*}

\section{Experimental Results}
\label{sec:exp}

\subsection{Datasets}
\label{sec:datasets}

\subsubsection{Training.}
For a fair comparison with~\cite{ fabbri2018enhancing, islam2020simultaneous, islam2020fast}, we pre-trained our model on ~1.3M ImageNet training split~\cite{russakovsky2015imagenet}, to reconstruct the input image (\ie, we set $\inputimg = \realcleanoutputimg$). 
The pre-trained \myalgoname is then fine-tuned for underwater image enhancement on the LSUI Train-L split proposed in~\cite{peng2023ushape-lsuidataset}. 

\subsubsection{Validation.}
We validate our method on six benchmark datasets to assess its generalization across diverse underwater conditions.
To perform a comparison between the enhanced image and the available ground truth, we considered the following full-reference datasets:
(i) the LSUI-L400 dataset~\cite{peng2023ushape-lsuidataset} comes with images featuring different water types, lighting conditions, and target categories\footnote{
The evaluation considers the Test-L 400 split proposed in\cite{peng2023ushape-lsuidataset}.};
(ii) the EUVP dataset\cite{islam2020fast} comprises 1970 validation image samples of varying quality;
and (iii) 
the UFO-120 dataset~\cite{islam2020simultaneous} contains 120 full-reference images collected from oceanic explorations across multiple locations and water types.

To assess enhanced image quality in a broader context, we further analyzed our model performance on the following non-reference datasets:
(i) the UCCS dataset~\cite{8949763} consists of 300 genuine underwater images captured across diverse marine organisms and environments, specifically designed to evaluate color cast correction in underwater image enhancement. (ii) the U45\cite{li2019fusion} and (iii) SQUID~\cite{berman2020underwater} datasets contain 45 and 57 raw underwater images showing severe color casts, low contrast, and haze.

\subsection{Metrics}
We followed recent works~\cite{li2019fusion,liu2022twin,peng2023ushape-lsuidataset,khan2024spectroformer}, and assessed our model performance considering the Peak Signal-to-Noise Ratio (PSNR), the Structural Similarity (SSIM)~\cite{wang2004ssim}, and the Learned Perceptual Image Patch Similarity (LPIPS)~\cite{zhang2018unreasonable} for full-reference datasets. 

For non-reference datasets, we considered the Underwater Color Image Quality Evaluation Metric (UCIQE)~\cite{7300447}, the Underwater Image Quality Measure (UIQM)~\cite{7305804}, the Natural Image Quality Evaluator (NIQE)~\cite{mittal2012making}, and the Inception Score (IS)~\cite{salimans2016improved}.

\subsection{Implementation Details}
We run the experimental evaluation with $\inputimg \in \mathbb{R}^{3\times H=256\times W=256}$.
$E$, having $N=5$ encoding blocks, yields $\mathbf{X}\in\mathbb{R}^{256\times 16\times 16}$.
$Q$ starts with $\beta_L=32$ to get $\mathbf{U} \in \mathbb{R}^{32\times 16 \times9 \times9}$.
From this, we get $32\times 9\times 9$ tensors each (of dimensionality $C_{\mathbf{U}}=16$) representing a capsule point of view.
In \myRbA{ }, we set $\alpha = 3$ to obtain $\hat{\mathbf{U}}\in \mathbb{R}^{64\times 32\times 9\times 9}$, where each vector obtained through the clusterization has $\beta_{L+1}=64$ dimensions. 
The normalization and following transposed convolution layers output $\mathbf{C} \in \mathbb{R}^{256\times 16\times 16}$.
We used the same settings as in~\cite{esser2021taming} for the pre-trained CNN considered in~(\ref{eq:lpips_loss}).
To optimize our loss function, we set $\delta = 10^{-6}$ following~\cite{esser2021taming}. 
We ran ImageNet pretraining for 25 epochs, with a batch size of $6$ using the Adam optimizer with a learning rate of $4.5e^{-6}$. 
Using the same optimization settings, we fine-tuned the resulting model on the LSUI Train-L dataset for 600 epochs using the adversarial strategy proposed in~\cite{esser2021taming}.
For both training processes, random cropping and horizontal flipping were applied.

\subsection{State-of-the-art Comparison}
\label{sec:exp:sota}

We compare the performance of our \myalgoname model with existing traditional methods like RGHS~\cite{huang2018rghs}, UDCP~\cite{drews2016udcp}, and UIBLA~\cite{peng2017uibla} as well as state-of-the-art machine learning-based works including UGAN~\cite{fabbri2018enhancing}, FUnIE-GAN~\cite{islam2020fast}, DeepSESR~\cite{islam2020simultaneous}, Cluie-Net~\cite{li2023cluienet}, TWIN~\cite{liu2022twin}, UShape-Transformer~\cite{peng2023ushape-lsuidataset}, and 
Spectroformer~\cite{khan2024spectroformer}.
We report on the results published in the corresponding papers or by running the publicly available codes.

\begin{table*}[t]
\centering
\scriptsize
\begin{tabularx}{\linewidth}{lcccccccccccc}
\toprule
                    & \multicolumn{4}{c}{UCCS} & \multicolumn{4}{c}{U45} & \multicolumn{4}{c}{SQUID} \\
                    \cmidrule(lr){2-5} \cmidrule(lr){6-9} \cmidrule(lr){10-13}
                    & UIQM $\uparrow$ & UCIQE $\uparrow$ & NIQE $\downarrow$ & IS $\uparrow$ & UIQM $\uparrow$ & UCIQE $\uparrow$ & NIQE $\downarrow$ & IS $\uparrow$ & UIQM $\uparrow$ & UCIQE $\uparrow$ & NIQE $\downarrow$ & IS $\uparrow$ \\
\midrule
RGHS~\cite{huang2018rghs}                & 2.97 & 0.55 & 5.14 & 2.23 & 2.57 & \tblfirst{0.62} & 4.34 & 2.21 & 1.46 & \tblsecond{0.56} & 9.25 & 2.02 \\
UDCP~\cite{drews2016udcp}                & 2.06 & 0.55 & 5.72 & 2.51 & 2.09 & 0.59 & 4.83 & 2.37 & 0.97 & \tblsecond{0.56} & 8.69 & 2.01 \\
UIBLA~\cite{peng2017uibla}               & 2.58 & 0.53 & 5.69 & 2.09 & 1.67 & 0.59 & 6.03 & 2.11 & 1.08 & 0.52 & 9.58 & 1.98 \\
UGAN~\cite{fabbri2018enhancing}          & 2.84 & 0.51 & 6.85 & 2.38 & 3.04 & 0.55 & 6.56 & 2.40 & 2.38 & 0.52 & 8.81 & 2.05 \\
Cluie-Net~\cite{li2023cluienet}          & 3.02 & 0.55 & 5.19 & 2.28 & 3.19 & 0.59 & 4.41 & 2.30 & 2.12 & 0.51 & 7.13 & 2.18 \\
TWIN~\cite{liu2022twin}                  & \tblfirst{3.23} & \tblfirst{0.59} & \tblfirst{4.45} & 2.22 & \tblfirst{3.36} & \tblfirst{0.62} & \tblfirst{4.16} & 2.30 & 2.31 & \tblfirst{0.57} & \tblfirst{6.44} & \tblfirst{2.21} \\
UShape-Transformer~\cite{peng2023ushape-lsuidataset} & 3.16 & \tblsecond{0.56} & 4.69 & \tblsecond{2.61} & 3.11 & 0.59 & 4.91 & 2.31 & 2.21 & 0.54 & 8.33 & 2.09 \\
Spectroformer~\cite{khan2024spectroformer} & \tblsecond{3.21} & 0.55 & 4.80 & 2.36 & 3.21 & \tblsecond{0.61} & \tblsecond{4.22} & \tblsecond{2.42} & \tblfirst{2.45} & \tblsecond{0.56} & \tblsecond{6.56} & 2.11 \\
\myalgoname                              & \tblsecond{3.21} & \tblsecond{0.56} & \tblsecond{4.65} & \tblfirst{2.65} & \tblsecond{3.23} & \tblsecond{0.61} & 4.29 & \tblfirst{2.44} & \tblsecond{2.44} & \tblfirst{0.57} & 6.58 & \tblsecond{2.19} \\
\bottomrule
\end{tabularx}
\caption{Quantitative comparison of \myalgoname and state-of-the-art methods across three non-reference datasets: UCCS, U45, and SQUID ($\uparrow$ higher is better, $\downarrow$ lower is better). For each metric/dataset the best method is in red, second best in blue.}
\label{tab:nonreference_sota_comparison}
\vspace{-1.5em}
\end{table*}

\subsubsection{Full-reference datasets.}
Table~\ref{tab:full_reference_sota} shows that across diverse underwater datasets, our method consistently outperforms existing underwater image enhancement approaches.
On the LSUI-L400 dataset, we achieve the best results considering all metrics, with a significant improvement in PSNR ($+1.47dB$) and SSIM ($+2\%$) compared to the best-performing method.
The evaluation of the EUVP dataset further highlights the capabilities of our approach with substantial improvements in SSIM ($+1\%$) and LPIPS ($-3\%$), showcasing its ability to restore image quality and perceptual fidelity. 
On the UFO-120 dataset, our method demonstrates notable improvements in PSNR ($+1dB$), SSIM ($+1\%$), and LPIPS ($-1\%$).
All such results substantiate the capabilities of our approach in leveraging a compressed latent space to precisely reconstruct the spatial relation between entities with great details under different water types, locations, lighting conditions, and multiple targets.

\subsubsection{Non-reference datasets.}
Results presented in Table~\ref{tab:nonreference_sota_comparison} show that our method demonstrates competitive performance across multiple non-reference datasets for underwater image enhancement.
This is particularly evident when we consider the IS metric --computed to evaluate how realistic the enhanced images are using a model pre-trained on natural images (\ie, ImageNet).
In such a case our method obtains the best overall results on all the datasets (\eg, $+0.04$ and $+0.02$ with respect to the best existing method on the UCCS and U45 datasets).
More in detail, on the UCCS dataset, we rank in second place in almost all metrics, except IS, for which we are at the top of the leaderboard.
A similar result is shown for the U45 dataset, where TWIN~\cite{liu2022twin} has the highest scores, and we ranked either second or third place, yet reaching the 1st place on the IS metric.
Similarly, on the SQUID dataset, our approach demonstrates competitive performance, achieving scores comparable to or higher than most existing methods across all metrics (\eg, $0.01$ difference between our method and the best performing one on UIQM and UCIQE metrics). 

\subsection{Ablation Study}
\label{sec:exp:ablation}
Through the ablation study, we want to answer different questions that would help us understand the importance of each proposed component of our architecture.

\begin{table}[t]
\centering
\scriptsize
\begin{tabularx}{\linewidth}{l C C C C}
    \toprule
    Method & PSNR $\uparrow$ & On-device Storage [MB] $\downarrow$ & Maximum Recording Duration [h] $\uparrow$ &Transmission time [s] $\downarrow$ \\
    \midrule
    Cluie-Net~\cite{li2023cluienet} & 18.71  & 1.57 & 0.39  & 0.01256 \\
    TWIN~\cite{liu2022twin} & 19.84 & 1.57  & 0.39 &  0.01256 \\
    UShape-T.~\cite{peng2023ushape-lsuidataset} & \tblsecond{23.02}  & \tblsecond{1.57} & \tblsecond{0.39} & \tblsecond{0.01256} \\
    Spectr.~\cite{khan2024spectroformer} & 20.09 &1.57 & 0.39 & 0.01256  \\
    \myalgoname & \tblfirst{24.49} & \tblfirst{0.52} &  \tblfirst{1.17} &   \tblfirst{0.00416} \\
    \bottomrule
\end{tabularx}
\caption{Enhancement performance and storage/transmission capabilities. Considering the real-world LSUI-L400 dataset, for a $256 \times 256$ input image, we computed the corresponding device storage space required for enhanced image generation and the related transmission time on a 1 GBbps bandwidth beacon. We also report on the maximum recording duration that would fit an off-the-shelf commercial device \protect\footnotemark ($\uparrow$ higher is better, $\downarrow$ lower is better).
}
\label{tab:psnr_vs_compression}
\vspace{-1em}

\end{table}

\subsubsection{How Efficient is the Encoder?}
Data storage and transmission are crucial for underwater data collection campaigns.
We designed our online encoder to store only the resulting latent image representation on the autonomous device.
This is the only information exploited by the offline dual-decoder module for full-size reconstruction and enhancement.
To assess the benefits of such a solution, we computed the results in Table~\ref{tab:psnr_vs_compression}.
Considering 30 samples to be enhanced, our method requires 15.6MB of storage and takes only 124.8ms to be transmitted through a 1Gbps bandwidth beacon.
All other methods considered for comparison required 47.1MB and 376ms, respectively.
These results demonstrate that our method, which takes approximatively 0.06 seconds to encode a single full-size input, offers a $3\times$ more efficient solution while also delivering the highest PSNR compared to state-of-the-art methods.

\begin{table}[t]
    \centering
    \scriptsize
    \begin{tabularx}{\linewidth}{l C C C}
        \toprule
        Method & EUVP & LSUI-L400 & UFO120 \\
        \midrule
        \myVQVAE~\cite{esser2021taming} & 20.58 & 19.36 & 19.62 \\
        \myalgoname\ w/o $D_S$ & \tblsecond{22.20} & \tblsecond{20.27} & \tblsecond{20.71} \\
        \myalgoname & \tblfirst{27.75} & \tblfirst{24.49} & \tblfirst{24.38} \\
        \bottomrule
    \end{tabularx}
    \caption{PSNR performance comparison between our three architecture variants and the \myVQVAE baseline.
    }
    \label{tab:ablation:vqvaecomparison}
\vspace{-1em}
\end{table}
\subsubsection{Is Capsule Clustering Good at Modeling the Latent Space?}
To explore the effectiveness of different latent information modeling methods, we replaced our capsule clustering procedure with the codebook learning process from VQ-VAE~\cite{esser2021taming}.
Table~\ref{tab:ablation:vqvaecomparison} shows that our novel capsule-based approach (\myalgoname w/o $D_S$, excluding the spatial decoder for equivalence) demonstrates a significant performance advantage over the VQ-VAE codebook variant, achieving an average improvement of more than$ 1.5$dB across the three full-reference datasets.
These results highlight the effectiveness of capsules, offering improved performance over discrete quantization methods that also suffer from non-differentiability issues.

\footnotetext{\url{https://planet-ocean.co.uk/surface-and-underwater-vehicles/}}

\begin{figure}[t]
  \centering
    \includegraphics[width=\linewidth]{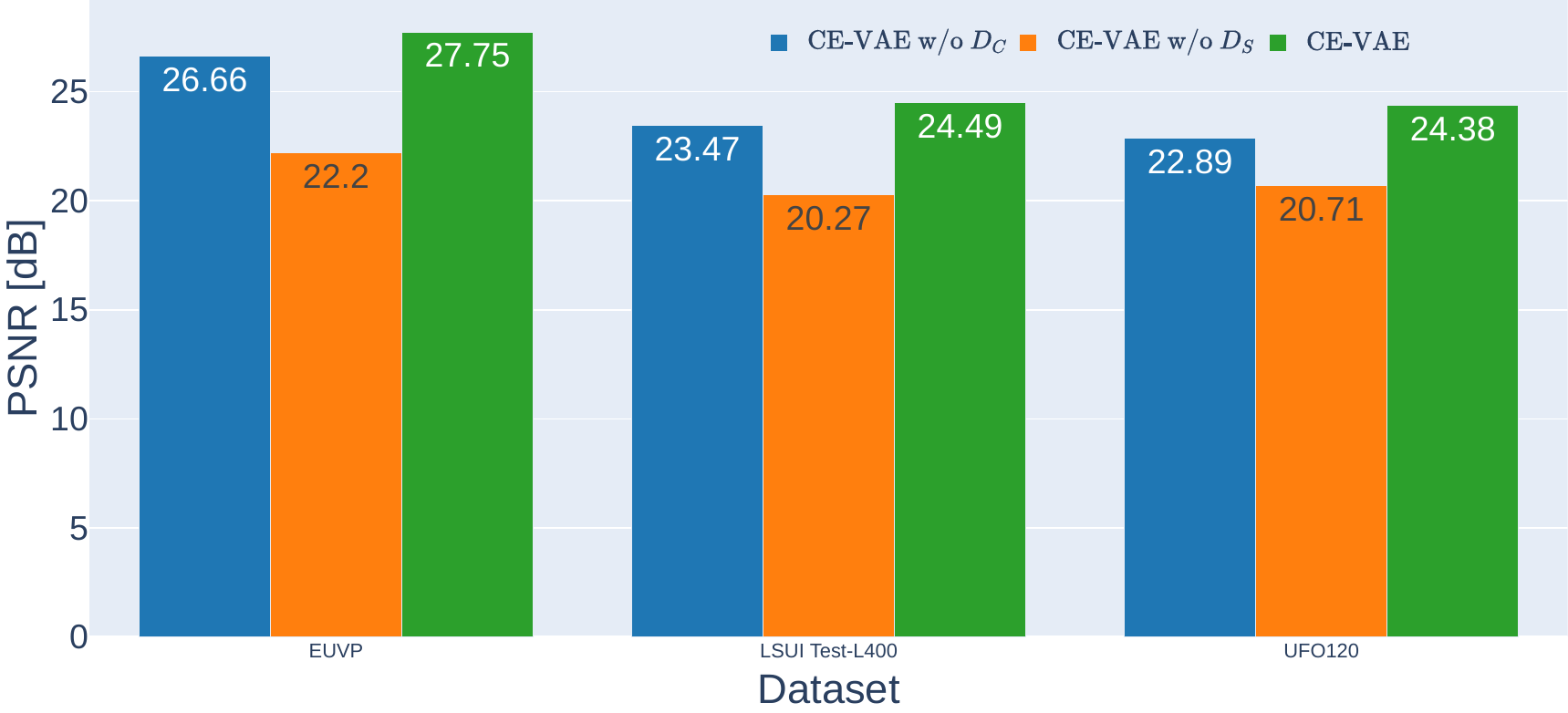}
  \caption{Analysis of the decoder components.
  Results are shown for our architecture (i) without the spatial decoder (\myalgoname w/o $D_S$), (ii) without the capsule decoder (\myalgoname w/o $D_c$), (iii) and for the complete \myalgoname.}
  \label{fig:ablation:caps_no_caps_skip_comparison}
\vspace{-1em}
\end{figure}

\begin{figure}[t]
  \centering
    \includegraphics[width=\linewidth]{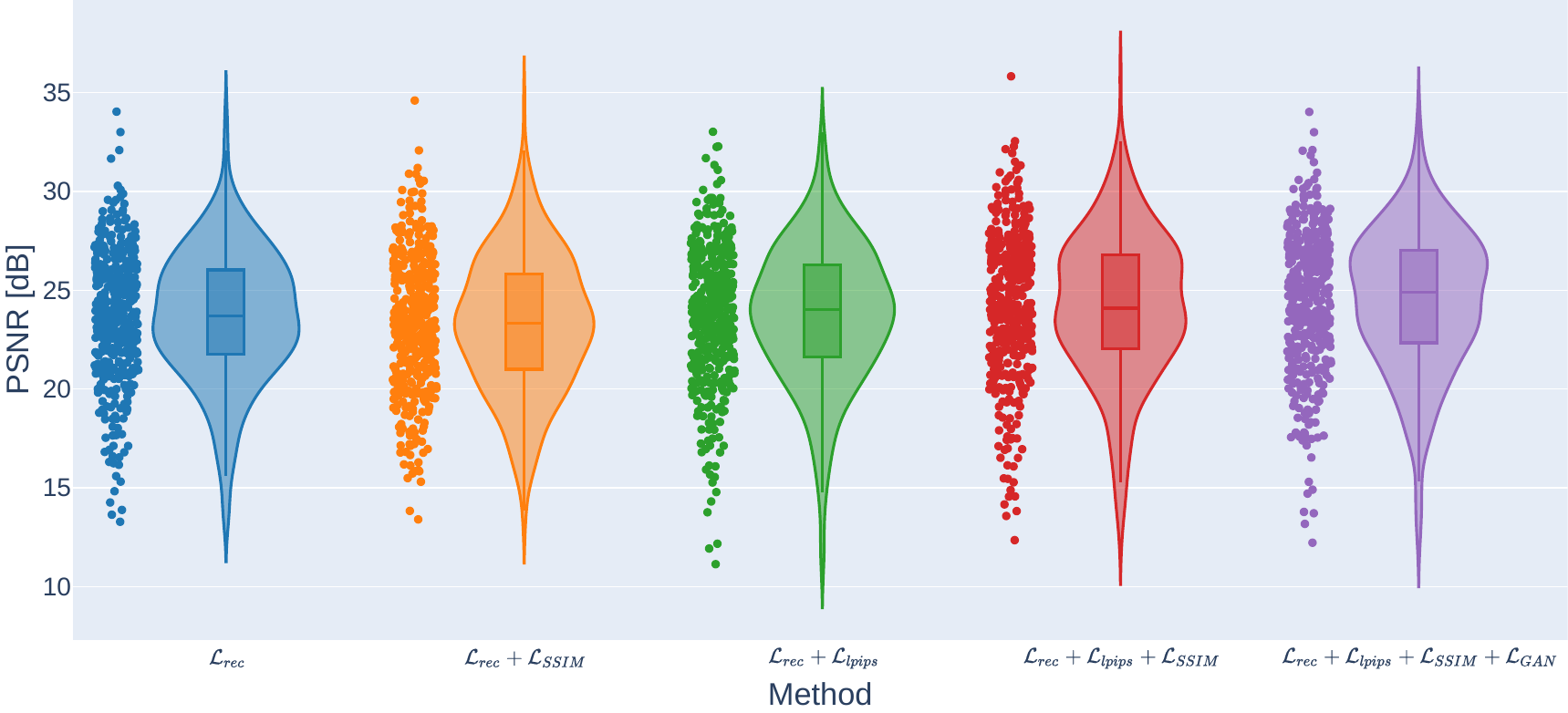}
    \vspace{-1em}
  \caption{Evaluation of the different loss components and their PSRN impact on the LSUI TestL-400 dataset.}
  \label{fig:ablation_losses}
  \vspace{-1.5em}
\end{figure}

\subsubsection{What is the Impact of Capsule and Spatial Decoders?}
\figurename~\ref{fig:ablation:caps_no_caps_skip_comparison}  illustrates the PSNR performance of our model with ablations of the capsule and spatial decoders.
Our \myalgoname model achieves superior PSNR scores across all datasets: $27.75$dB on EUVP, $24.49$dB on LSUI Test-L400, and $24.38$dB on UFO120.
While the capsule decoder alone (\myalgoname w/o $D_S$) yields results on par with existing methods (\eg, $22.2$dB on EUVP), the spatial decoder (\myalgoname w/o $D_\mathbf{C}$) demonstrates significantly higher performance (\eg, $26.66$dB on EUVP), demonstrating its importance for pixel-level reconstruction.
These results substantiate our intuition on designing these two encoders that, leveraging their complementary strengths, effectively model entity-related and spatially precise details for image enhancement.

\begin{figure*}[b]
  \centering
  \includegraphics[width=0.92\linewidth]{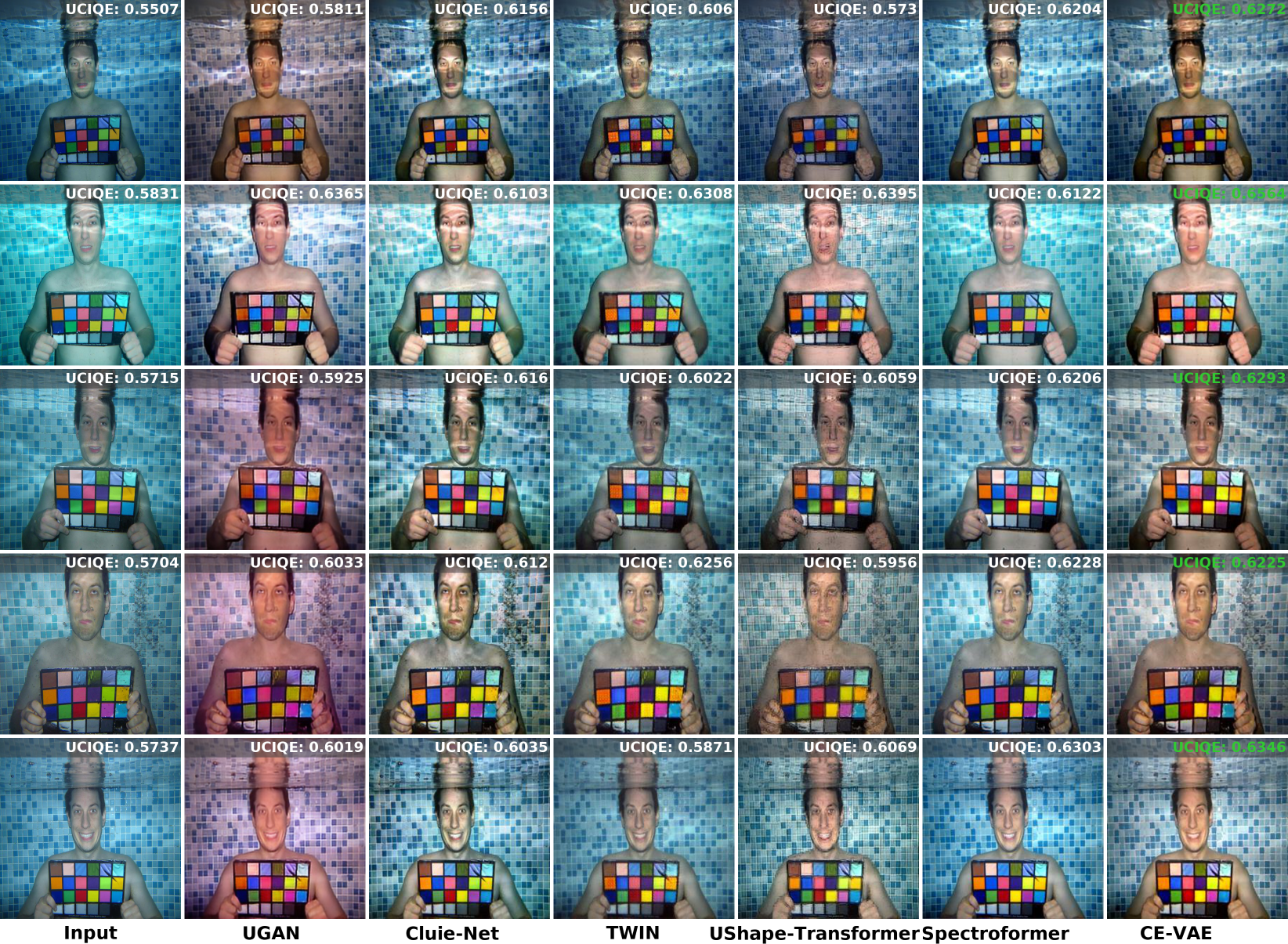}
  \caption{Enhanced images comparison on the Color-Check7 dataset.} 
  \label{fig:sota_qualitative_comparison_color_check7}
  \vspace{-1em}
\end{figure*}

\subsubsection{How Relevant Are the Loss Terms?}
\figurename~\ref{fig:ablation_losses} shows the results obtained considering the LSUI Test-L 400 dataset by our approach when loss components are turned on and off (the analysis on the other validation datasets is in the supplementary).
The performance shows a PSNR increase as more loss terms are included. 
Notably, including the adversarial loss (\ie, $\loss_{GAN}$) concentrates the distribution of PSNR values around the mean, highlighting the robustness of the method towards edge cases.
This is likely to be due to the GAN discriminator role that might prevent the generation of unfeasible results containing hallucinated artifacts or values that are close to the real one in the RGB space, but result in different semantics.

\subsection{Qualitative Analysis}
\label{sec:qualitative_analysis}

To showcase the robustness and accuracy of our method for underwater color correction, we report on the qualitative performance comparison between our method and existing solutions on the Color-Checker7 dataset~\cite{ancuti2018colorbalance}.
The dataset contains 7 underwater images captured with different cameras, each including a Macbeth Color Checker.
This allows us to assess our method's performance across diverse imaging devices and its ability to accurately restore true colors.
\figurename~\ref{fig:sota_qualitative_comparison_color_check7} shows that our method provides a neat color restoration for all items in the Macbeth Color Checker while also yielding a realistic color for the skin.

\section{Conclusions}
\label{sec:conclusion}
We introduced a novel architecture for underwater image enhancement featuring a highly efficient encoder module that yields $3\times$ compression efficiency compared to existing approaches while running online. 
To decode the encoded, compressed, representation we proposed a dual-decoder module that leverages spatial and entity-level information (captured by a capsule layer) to precisely reconstruct a full-size enhanced version of the input.
Our approach demonstrates superior image quality across six benchmark datasets while potentially facilitating longer missions and more comprehensive data collection.

{\small
\bibliographystyle{ieee_fullname}
\bibliography{manuscript}
}
\end{document}


\title{CE-VAE: Capsule Enhanced Variational AutoEncoder for Underwater Image Enhancement: tech report}

\author{Rita Pucci\\
Natuarlis Biodivercsity Center\\
Leiden, Netherlands (NL)\\
{\tt\small rita.pucci@naturalis.nl}
\and
Niki Martinel\\
Department of Mathematics, Computer Science and Physics, University of Udine\\
Udine, Italy (IT)\\
{\tt\small niki.martinel@uniud.it}
}
\maketitle

\section{Introduction}
In this tech report, we present and describe additional studies performed during the research for the ablation study. In particular, we report (i) the impact of the loss components described in the paper on EUVP and UFO120 datasets; and (ii) we show additional results obtained on the UFO120 and LSUI Test-L400 datasets.

\section{Ablation Study}
\label{sec:exp:ablation}
In the paper, we presented the ablation study, which aims to address various questions that will help us comprehend the significance of each suggested component in our architecture. We showed the impact of the different components of the loss presenting results with the LSUI Test-L 400 dataset. The report presents additional results for the EUVP and UFO120 datasets. 

\begin{figure}
  \centering
\begin{subfigure}{\linewidth}
    \includegraphics[width=\linewidth]{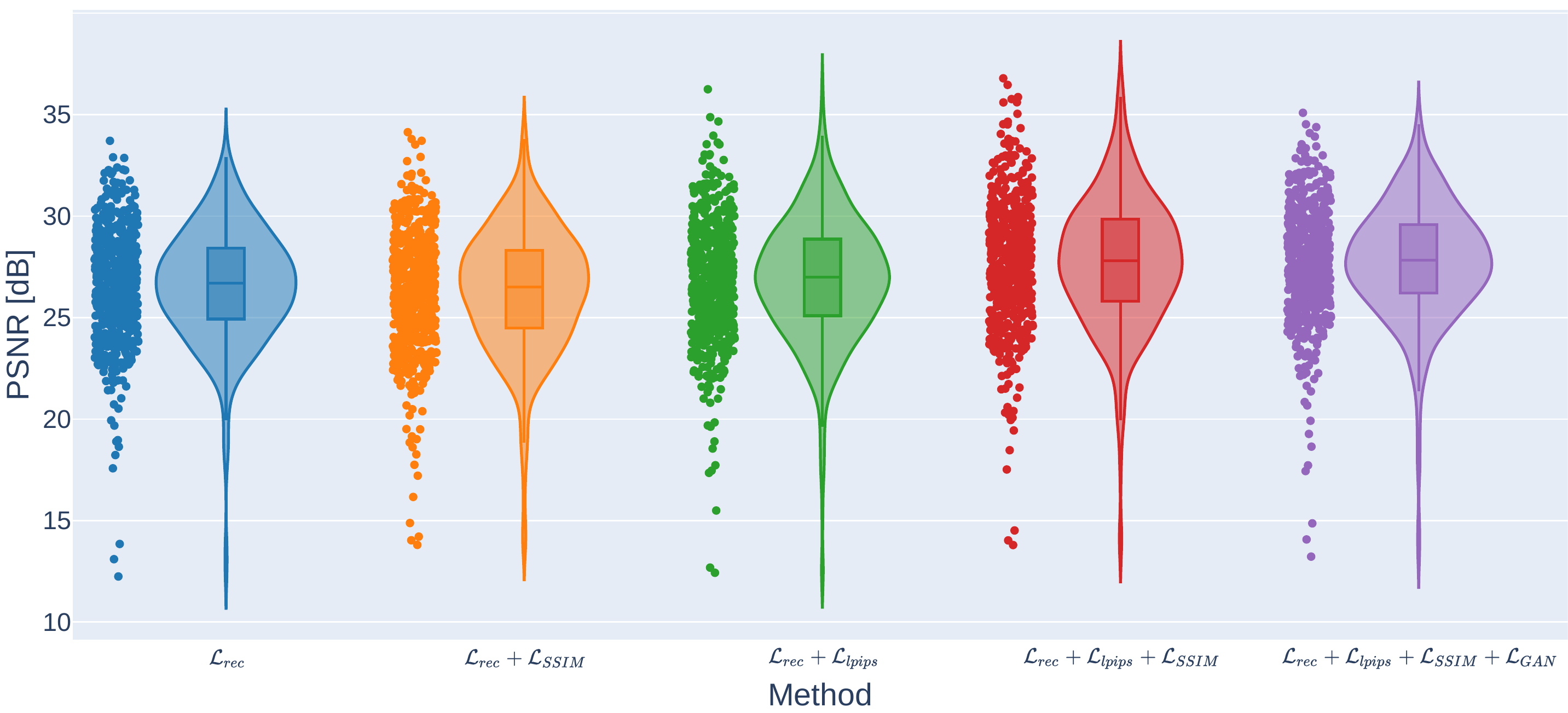}
    \caption{EUVP}
    \label{fig:ablation:losses:euvp}
\end{subfigure}
\hspace{2em}
\begin{subfigure}{\linewidth}
    \includegraphics[width=\linewidth]{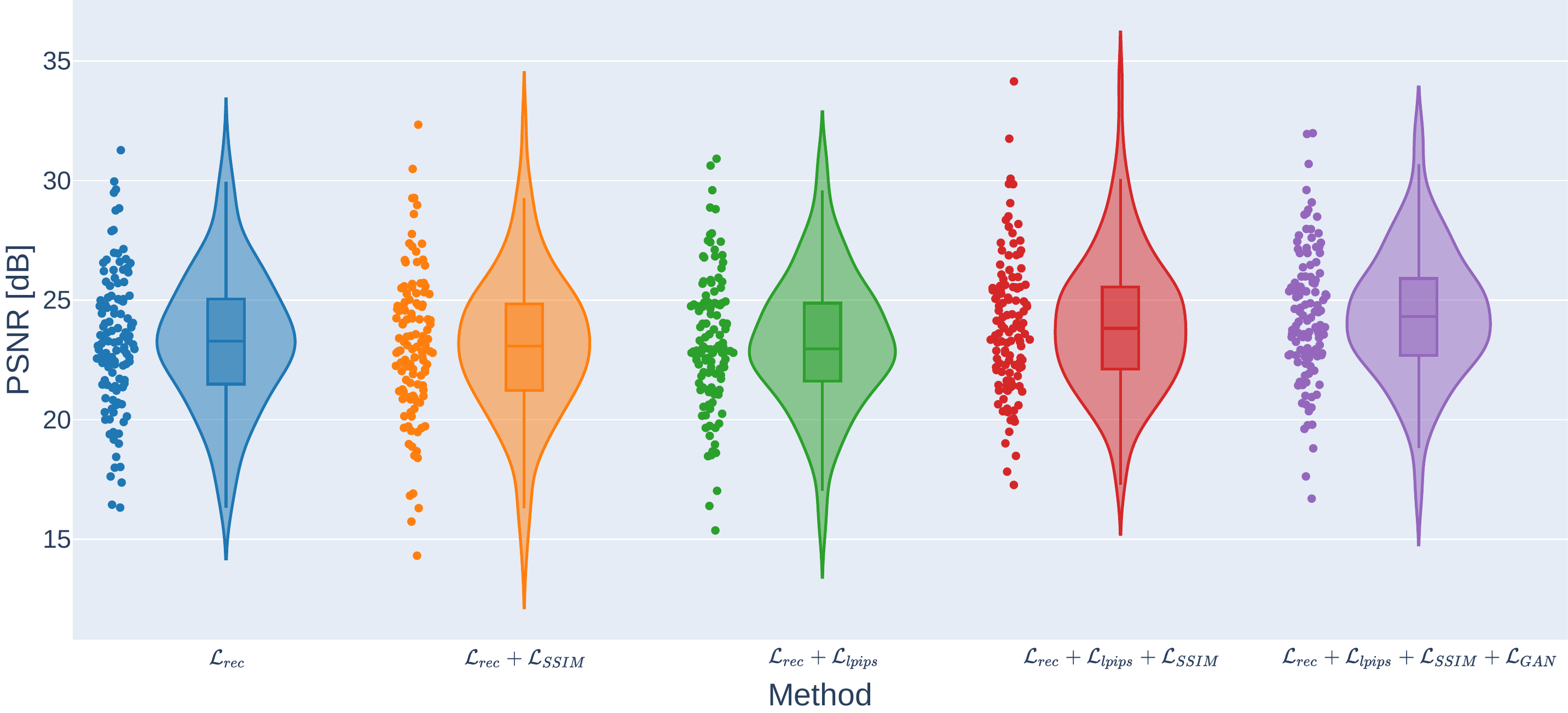}
    \caption{UFO-120}
    \label{fig:ablation:losses:ufo}
\end{subfigure}
  \caption{Evaluation of the different loss components and their PSRN impact on the  (\subref{fig:ablation:losses:euvp}) the EUVP dataset, and (\subref{fig:ablation:losses:ufo}) the UFO-120 dataset.}
  \label{fig:ablation_losses}
\end{figure}

\begin{figure*}
  \centering
  \includegraphics[width=0.9\linewidth]{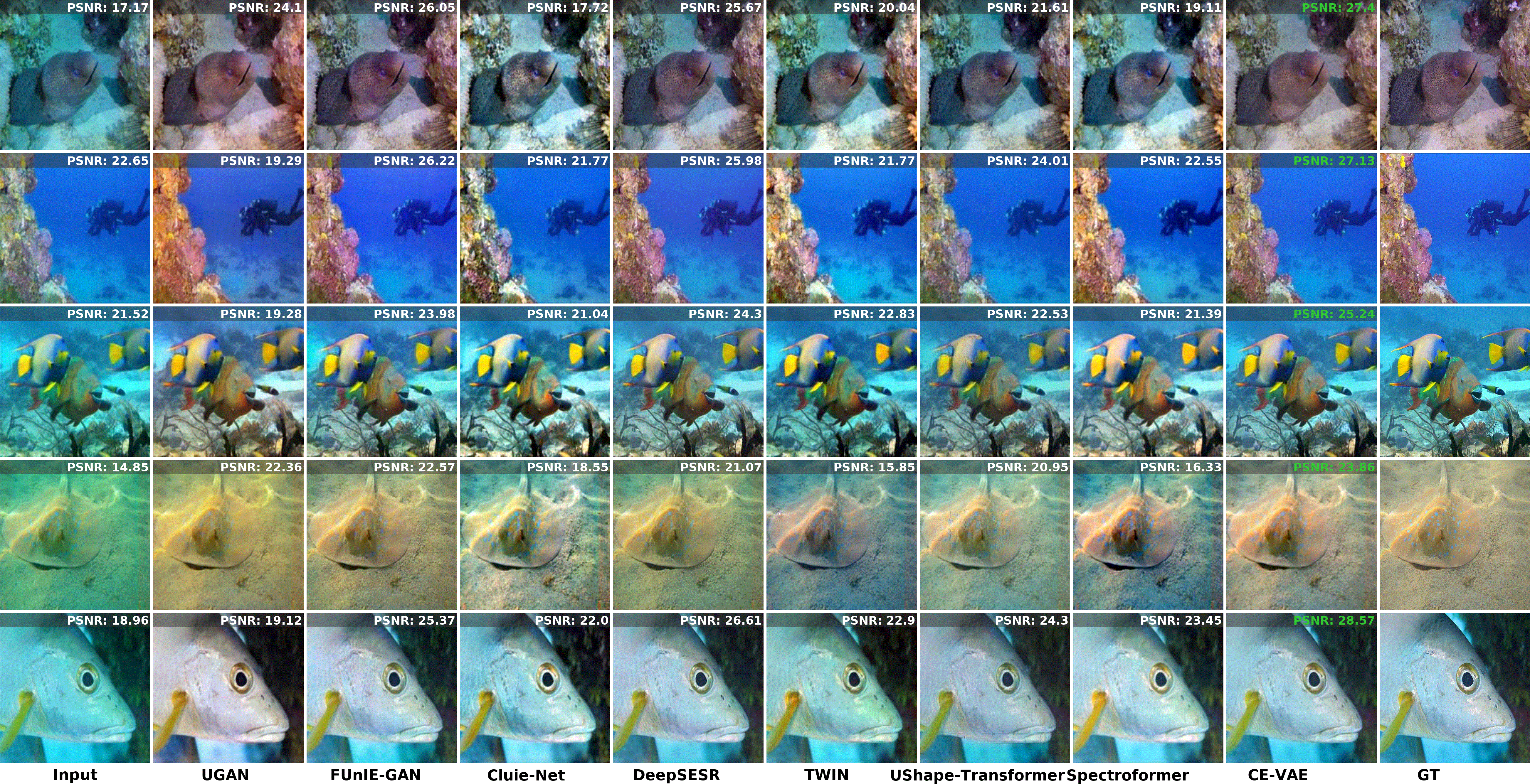}
  \caption{Enhanced images comparison on the UFO120 dataset.} 
  \label{fig:sota_qualitative_comparison_ufo}
\end{figure*}
\begin{figure*}
  \centering
  \includegraphics[width=0.9\linewidth]{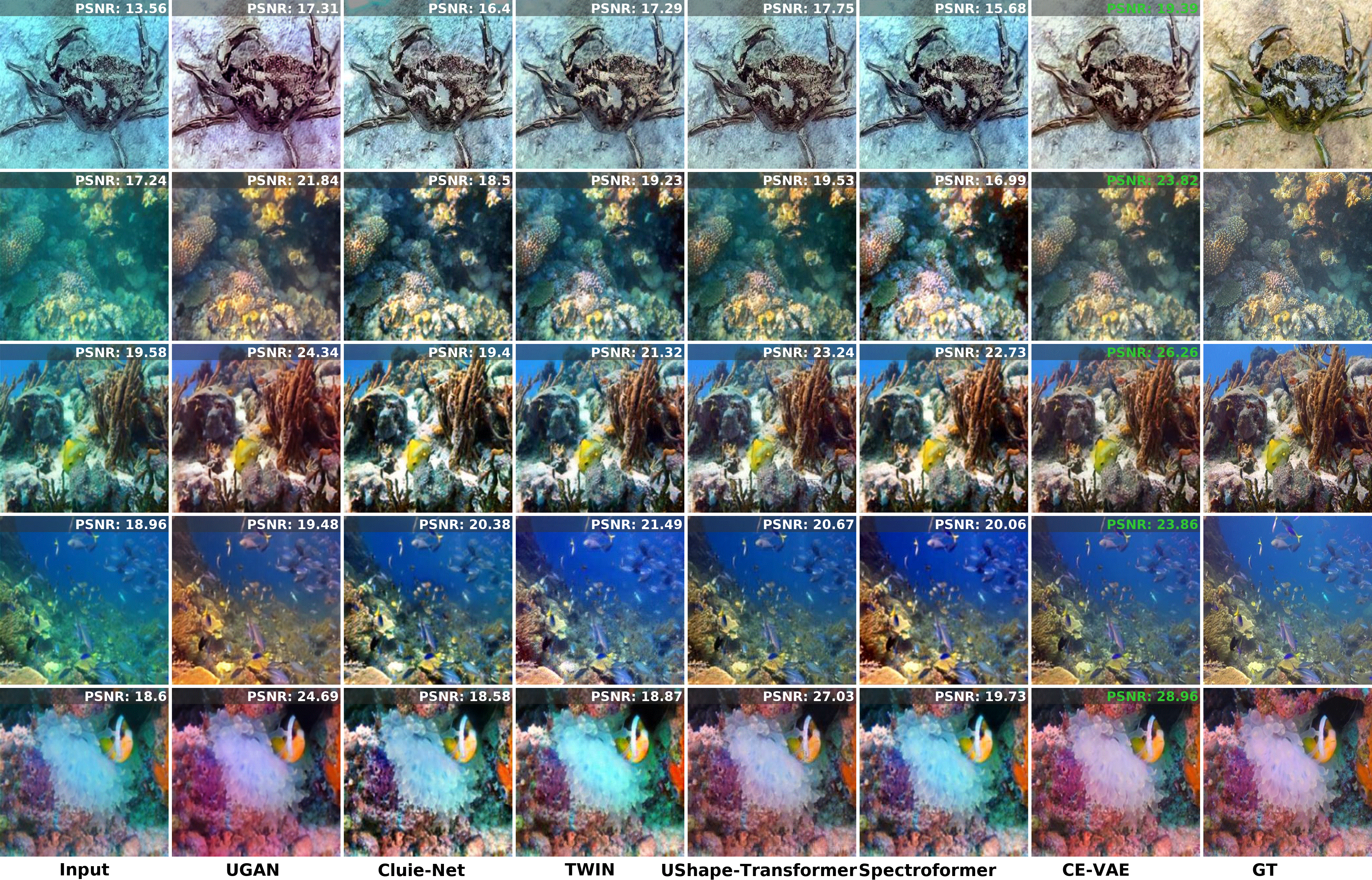}
  \caption{Enhanced images comparison on the LSUI Test-L400 dataset.} 
  \label{fig:sota_qualitative_comparison_lsui}
\end{figure*}

The \figurename~\ref{fig:ablation_losses} shows the results obtained by our approach when loss components are turned on and off. On the two considered datasets, the performance exhibits similar behaviour to the one presented in the paper with the LSUI Test-L 400 dataset. The average PSNR increases as more loss terms are included and by including the adversarial loss (\ie, $\loss_{GAN}$), concentrates the distribution of PSNR values more around the mean, indicating the robustness of the method towards edge cases. As mentioned in the paper, this could be attributed to the GAN discriminator, which helps prevent the generation of unrealistic artifacts or values that might look real in terms of color but have different semantics.

\section{Qualitative analysis}
The results from the Macbeth Color Checker, as presented in the paper, demonstrate the superior performance of the CE-VAE model compared to the state-of-the-art methods. Our model consistently produces clean and realistic colors, surpassing existing solutions. The report highlights the effectiveness of our method for underwater color correction and includes additional results for qualitative performance comparisons on the UFO120 and LSUI Test-L400 datasets. Fig.~\figurename~\ref{fig:sota_qualitative_comparison_ufo} and ~\figurename~\ref{fig:sota_qualitative_comparison_lsui} illustrate the detailed color reconstruction and comparison with ground truth colors, as well as the comparison with results obtained using state-of-the-art methods. The PSNR value is displayed in the upper-right corner of each generated image. Overall, the CE-VAE model outperforms all other models on both datasets.